\documentclass[oneside,11pt]{article}

\usepackage{geometry}

\usepackage{amssymb}

\usepackage{mathtools}
\usepackage{amsthm}

\usepackage[capitalize,noabbrev]{cleveref}
\usepackage{amsmath}

\newcommand{\R}{\mathbb{R}}

\newcommand{\N}{\mathbb{N}}

\newcommand{\numb}{\boldsymbol{r}}

\newcommand{\ar}{\mathcal{A}}

\newcommand{\A}{\boldsymbol{z}}
\newcommand{\cH}{\mathcal{H}}
\newcommand{\cHP}{\Lambda}
\newcommand{\act}{\phi}
\newcommand{\Act}{\boldsymbol{\Phi}}

\newcommand{\gputph}{\textbf{gp$/$ph$\,$}}
\newcommand{\rank}{\mathrm{Rank}}

\newcommand{\afix}{\mathrm{StblFix}}
\newcommand{\fix}{\mathrm{Fix}}

\newcommand{\poly}{\mathcal{R}}

\newcommand{\hi}{\boldsymbol{h}}

\newcommand{\bx}{\boldsymbol{x}}

\newcommand{\by}{\boldsymbol{y}}
\newcommand{\bv}{\boldsymbol{v}}
\newcommand{\bu}{\boldsymbol{u}}
\newcommand{\bq}{\boldsymbol{q}}
\newcommand{\br}{\boldsymbol{r}}

\newcommand{\conn}{\mathcal{C}}

\newcommand{\bz}{\boldsymbol{z}}
\newcommand{\ba}{\boldsymbol{a}}
\newcommand{\bc}{\boldsymbol{c}}
\newcommand{\bW}{\boldsymbol{W}}
\newcommand{\bV}{\boldsymbol{V}}
\newcommand{\bA}{\boldsymbol{A}}

\newcommand{\bnu}{\boldsymbol{\nu}}
\newcommand{\bom}{\boldsymbol{\omega}}

\newcommand{\bsig}{\boldsymbol{\Sigma}}

\newcommand{\bD}{\boldsymbol{D}}
\newcommand{\Jac}{\boldsymbol{J}}

\newcommand{\conv}{\mathrm{cv}}
\newcommand{\inn}{\mathrm{int}}

\newtheorem{theorem}{Theorem}[section]
\newtheorem{proposition}[theorem]{Proposition}

\newtheorem{lemma}[theorem]{Lemma}

\theoremstyle{definition}
\newtheorem{definition}[theorem]{Definition}

\begin{document}

\title{Hyperplane Arrangements and Fixed Points in Iterated PWL Neural Networks  

}
	\author{
 	Hans-Peter Beise\\
  	Department of Computer Science\\
 	Trier University of Applied Sciences\\
 	}

\maketitle     

\begin{abstract}
We leverage the framework of hyperplane arrangements to analyze potential regions of (stable) fixed points.  Expanding on concepts from \cite{pascanu2013number, montufar2014number,serra2018bounding}, we provide an upper bound on the number of fixed points for multi-layer neural networks equipped with piecewise linear (PWL) activation functions with arbitrary many linear pieces. The theoretical optimality of the exponential growth in the number of layers of the latter bound is shown. Specifically, we also derive a sharper upper bound on the number of stable fixed points for one-hidden-layer networks with hard tanh activation.

\end{abstract}








\section{Introduction}\label{sec:Introduction}

Inspired by theoretical investigations on network complexity \cite{raghu2016expressive, pascanu2013number, montufar2014number, serra2018bounding, montufar2022sharp, xiong2020number}, we utilize hyperplane arrangements to derive upper bounds on the number of fixed points. We establish an upper bound on the number of fixed points for multi-layer neural networks with piecewise linear activation \textbf{(PWL)} functions. Our result includes networks with PWL activation functions with several linear pieces, distinguishing our work from related investigations. 
Using a saw-tooth-like construction of a neural network \cite{telgarsky2016benefits}, we demonstrate that the exponential growth in the number of layers in our bound is theoretically optimal. 
Combining the analysis of the spectral norm of Jacobian matrices with characteristics of certain activation regions, we show a dedicated upper bound on the number of stable fixed points for one-hidden-layer neural networks with hard tanh activation.




\section{Prelimaries}\label{sec:prem}

We introduce some basic notation and terminology employed throughout.

Vectors and matrixes are denoted by boldface letters, and $\bx^{(j)}$ signifies the $j$-th component of vector $\bx$, and $\Vert \bx\Vert$ means the Euclidean norm. For a matrix $\bA$, $\Vert \bA\Vert :=\max_{\Vert \bx\Vert=1}\Vert\bA\bx\Vert$ is the spectral norm. For a mapping $f:A\rightarrow B$ and $M\subset A$, we use $f\vert_M$ to denote the restriction of $f$ to the subdomain $M$. The composition of functions is denoted by $\circ$. The convex hull of a set $M\subset \R^n$ is denoted by $\conv(M)$, and the \textbf{closure} of $M$, which encompasses the points in $M$ along with all limit points, is denoted by $\overline{M}$. The set of \textbf{interior points} of $M$ is denoted by $\inn(M)$, and its \textbf{boundary} is defined by $\partial M=\overline{M}\setminus \inn(M)$. By $\vert M \vert$ we denote the number of elements in a finite set $M$. Additional notations are introduced as needed in subsequent sections.

Let $f: \mathbb{R}^d \rightarrow \mathbb{R}^d$ be a mapping. A vector $\bx^* \in \mathbb{R}^d$ is called a \textbf{fixed point} (of $f$) if $f(\bx^*) = \bx^*$. The set of all fixed points of $f$ is denoted by $\fix(f)$.
A vector $\bx^* \in \mathbb{R}^d$ is called an \textbf{attractive fixed point} if there exists a non-empty open neighborhood $\mathcal{U}$ of $\bx^*$ such that for all $\bx \in \mathcal{U}$, $f^{\circ k}(\bx) \rightarrow \bx^*$ as $k \rightarrow \infty$, where $f^{\circ k}$ denotes the $k$-fold iteration of $f$.
If $f$ is differentiable in a neighborhood of a fixed point $\bx^*$, then we say that $\bx^*$ is a \textbf{stable fixed point} if the spectral norm of the Jacobian matrix of $f$ at $\bx^*$ is strictly less than 1. The set of all stable fixed points of $f$ is denoted by $\afix(f)$. Note that $\afix(f)$ consists of isolated fixed points, and all stable fixed points are attractive.
It is important to mention that the study of stable fixed points, as introduced earlier, excludes possible fixed points in subdomains where $f$ is not differentiable. However, in common neural networks, such subdomains typically constitute a set of zero Lebesgue measure.

 We consider networks functions $f: \R^{d} \rightarrow \R^{d}$ (autoencoders), layer-wise defined:
\begin{equation}\label{networkFunLayered}
	f = \hi_L \circ \hi_{L-1} \circ \ldots \circ \hi_1,
\end{equation} 
where each layer function $\hi_j: \R^{n_{j-1}} \rightarrow \R^{n_j}$ is defined as $\hi_j(\bx) = \Act(\bW_j\bx + \bu_j)$ with $\bW_j \in \R^{n_{j}\times n_{j-1}}$, $\bu_j \in \R^{n_j}$, for $j = 1, \ldots, L$, and $d = n_1 = n_L$. The activation $\Act:\R^n\rightarrow \R^n$ is component wise defined by $\Act=(\act,\ldots,\act)^T$, where $\act:\R\rightarrow\R$ is a piecewise linear activation \textbf{(PWL)}, which is always assumed to be continuous.

In particular, we are also interested in networks $f: \R^{d} \rightarrow \R^{d}$, taking the following form:
\begin{equation}\label{networkFun00}
	f(\bx) = \bW\, \Act\left(\bV\bx + \bu\right) + \bz,
\end{equation}
where $\bV, \bW^T \in \R^{n \times d}$ and $\bu \in \R^n$, $\bz \in \R^d$. 
A simplified version of (\ref{networkFun00}) writes:
\begin{equation}\label{networkFun1}
	f(\bx) = \bW\, \Act\left(\bW^T\bx\right).
\end{equation}


\section{Arrangements of Parallel and Non-Parallel Hyperplanes}\label{sec_hyperplane}

We utilize hyperplane arrangements to establish upper bounds on the number of fixed points. To facilitate our discussion, we introduce some key notions and results on hyperplane arrangements, and refer to \cite{rstandely2004} for a detailed exploration of the topic.

For a set of hyperplanes $\cH := \{H_1, \ldots, H_n\}$, referred to as a \textbf{hyperplane arrangement}, in $\R^d$, let $\poly(\cH)$ denote the connected components of $\R^d \setminus \left(H_1 \cup \ldots \cup H_n\right)$. We will refer to the elements of $\poly(\cH)$ as \textbf{regions}, and define $ \numb_d(\cH) := \vert\poly(\cH)\vert$.

Zaslavsky's theorem \cite{zaslavsky1975facing}, see also [Proposition 2.4] \cite{rstandely2004}, states that the number of regions of an arrangement of $n$ hyperplanes in $\R^d$ is upper bounded by 
\begin{equation}\label{zaslavsky_upper}
    \numb_d(\cH) \leq \sum_{j=0}^d\binom{n}{j}.
\end{equation}
This upper bound is attained if, and only if, the arrangement is in a generic configuration called \textbf{general position}, i.e. if for every sub-arrangement $\{H_1, \ldots, H_r\} \subset \cH$, the dimension of $\bigcap_{j=1}^r H_j$ equals $d-r$ for $r<d$ and is zero for $r \geq d$, respectively. It is important to emphasize that general position represents the prevalent scenario. A technically involved generalisation to arrangements possibly not in general position  is given in \cite{xiong2020number} for the sake of analyzing regions of convolutional neural networks. Compared to the latter, our study can be reduced to a simplified scenario, allowing us to derive more straightforward formulas.

In the context of neural networks with PWL activation, the following type of hyperplane arrangement naturally comes into play, cf. \cref{sec:Appendix_HyperProof}.

\begin{definition}\label{def:paraHyperArr}
    For $n, k \in \N$, let $H_{j,l}$ be hyperplanes in $\R^d$, with $j\in\{1, \ldots, k\}$ and $l\in \{1, \ldots, n\}$ such that the following hold. For fixed $l$, all hyperplanes $H_{j,l}$ are \textbf{parallel but not equal}. For pairwise distinct $l_1, \ldots, l_n$ and arbitrary $j_1, \ldots, j_n$ in $\{1, \ldots, k\}$, the hyperplanes $H_{j_1,l_1}, \ldots, H_{j_n,l_n}$ are in general position. We will refer to this property as \textbf{general position modulo parallel hyperplanes (\gputph)}. The set of hyperplane arrangements that satisfy the above conditions is denoted by $\cHP_d(n,k)$.
\end{definition}

Due to the property of being in general position modulo parallel hyperplanes (\gputph), every arrangement in $\cHP_d(n,k)$ exhibits an equal number of regions, as elaborated in \cref{lemma_sameNumerRegionInHs}. This quantity is denoted by 
\begin{equation}\label{numb_paraHyper}
    \numb_d(n,k):=\vert\poly\left(\cH \right)\vert, \ \cH \in \cHP_d(n,k),
\end{equation}
where we abbreviate $\numb_d(n,1)=\numb_d(n)$.

\begin{proposition}\label{th:numb_explicit}
For $n,k\in\N$, the number of regions for every $\cH\in \cHP(n,k)$ is given by
\begin{equation}\label{numbHyper}
   \numb_d(n,k)= (k+1)\sum_{j=0}^{d-1}\binom{n-1}{j} k^j+k^{d}\sum_{j=d}^{n-1}\binom{j-1}{d-1}.
\end{equation}
\end{proposition}
\begin{proof}
    The proof relies on the recursive equation $\numb_d(n,k)=\numb_d(n-1,k)+k\numb_{d-1}(n-1,k)$, for $n,d>1$, which is found to unfold to  (\ref{numbHyper}) by \cref{lemm:unfold_rec}. Details are given in \cref{sec:Appendix_HyperProof}.
\end{proof}

We are particularly interested in specific regions of arrangements $\cH \in \cHP_d(n,2)$.

\begin{definition}\label{def:hyperPlusMins}
Consider $\bom_1, \ldots, \bom_n \in \mathbb{R}^d \setminus \{0\}$ and $a_j < b_j$ for $j \in \{1, \ldots, n\}$. Define the hyperplanes in the arrangement $\cH := \{H_{1,j}, H_{2,j}: j=1, \ldots, n\}$ by $H_{1,j} := \{\bx \in \mathbb{R}^d: \bom_j^T\bx = b_j\}$ and $H_{2,j} := \{\bx \in \mathbb{R}^d: \bom_j^T\bx = a_j\}$ and assume that $\cH\in \cHP_d(n,2)$. Let
\begin{eqnarray}
    \bom_j^T\bx &\leq a_j \label{cond_pmH1}, \\ 
    \bom_j^T\bx &\geq b_j\label{cond_pmH2},
\end{eqnarray}
and define $\poly^\pm(\cH) \subset \poly(\cH)$ as those regions $C \in \poly(\cH)$ where, for all $\bx \in C$, either (\ref{cond_pmH1}) or (\ref{cond_pmH2}) holds for all $j \in \{1, \ldots, n\}$.
\end{definition}

To briefly motivate \cref{def:hyperPlusMins}, let $\act$ be the sigmoid function  and $c > 0$. We can determine $a > 0$ such that $\act^\prime(x) \leq c$ on $(-\infty, -a]$ and $[a, \infty)$, while $\act^\prime(x) \geq c$ on $[-a, a]$. If $\act$ is applied in (\ref{networkFun00}), arrangements as defined in \cref{def:hyperPlusMins} become of interest, cf. \cref{sec:nonlinearFP}.

\begin{proposition}\label{th:upperLowerPMHyper}
The number of regions in $\vert \poly^\pm(\cH)\vert $ for an arrangement of hyperplanes $\cH$ as in \cref{def:hyperPlusMins}, is upper bounded by
\begin{equation}\label{upperLowerPMHyper}
   \vert \poly^\pm(\cH)\vert \leq \numb_d(n,1)=\sum_{j=0}^d\binom{n}{j} .
\end{equation}
\end{proposition}
Elementary examples reveal that both, equality and inequality can hold in (\ref{upperLowerPMHyper}), even though \gputph holds by assumption. 
\begin{proof}(\cref{th:upperLowerPMHyper})
   All arrangements within this proof are of the kind specified in \cref{def:hyperPlusMins}.
   The initial bounds $\vert \poly^\pm (\cH)\vert\leq n+1=:r(n,1)$, $\vert \poly^\pm (\cH)\vert2=:r(1,d)$ for all $\cH\in\cHP_1(n,2)$, $\cH\in\cHP_{d}(1,2)$, respectively, are obvious. Assuming $r(n-1,d)$ and $r(n-1,d-1)$ upper bound the number of all regions in $\poly^\pm$ for arrangements in $\cHP_d(n-1,2)$, $\cHP_{d-1}(n-1,2)$,  respectively, it is shown in \cref{lemm:upperBoundHpyer_recusion} that
       $\vert \poly^\pm(\cH)\vert\leq r(n-1,d)+r(n-1,d-1)=:r(n,d).$
  \cref{lemm:unfold_rec} then yields (\ref{upperLowerPMHyper}). 
\end{proof}

\section{Linear Regions and Fixed Points of Multi-Layer Neural Networks}\label{sec:actregion}

We follow previous works that investigate linear regions of networks with PWL activation functions \cite{pascanu2013number,montufar2014number,serra2018bounding,xiong2020number,montufar2022sharp}, utilizing this frameworks to derive an upper bound on fixed points for multi-layer networks. We first need to specify some terminology, wherein we interchangeably use the term \textbf{linear} with the mathematical term \textbf{affine}.

\begin{definition}\label{def_pwActivation}
A continuous function $\act:\R\rightarrow \R$ is said to by \textbf{piecewise linear} (\textbf{PWL}), if there are open, disjoint intervals $I_1,\ldots,I_k$ with $\R=\bigcup_{j=1}^k \overline{I_j}$ such that $\act\vert_I$ is a linear for $I\in \{I_1,\ldots,I_k\}$. We call $\ar(\act):=\{I_1,\ldots,I_k\}$ the \textbf{linear regions} of $f$, where we assume that the $I\in \ar(\act)$ are maximal in the sense that $\act\vert_Z$ is not linear for $I\subset Z$ when $Z\neq I$. 
\end{definition}

\begin{definition}\label{def_PWL_net}
Let $f$ be a multi-layer network as in (\ref{networkFunLayered}), endowed with a PWL activation function $\act$. A \textbf{linear region} of $f$ is a set $C \subset \mathbb{R}^{d}$ consisting of all points $\bx$ that share the property that for all layers $l=1,\ldots,L$ and every component (neuron) $j=1,\ldots,n_l$ in those layers, there exists $I(j,l,C) \in \ar(\act)$ such that for all $\bx\in C$
\begin{equation*}
  (\bW_l\,(\hi_{l-1} \circ \ldots \circ \hi_1(\bx))+\bu_l)^{(j)} \in I(j,l,C).  
\end{equation*}
The set of all linear regions is denoted by $\ar(f)$.
\end{definition}

As in Theorem 2 of \cite{raghu2016expressive}, it follows that $\ar(f)$ partitions the input space into convex sets. 

\begin{lemma}\label{fixedPointsPWL}
Let $f$ be a network function as in (\ref{networkFunLayered}) endowed with a PWL activation function $\act:\R\rightarrow \R$, then $C\cap\fix(f)$ is convex for all  $C\in \ar(f)$.
\end{lemma}

\cref{fixedPointsPWL} and the arguments commonly utilized to derive upper bounds on activations regions \cite{montufar2017notes,serra2018bounding}, now enable to upper bound the number of components of $\fix(f)$. In contrast to previous works in that direction, our analysis includes explicit bounds for PWL activations with several linear pieces. 

We recall that any $M\subset \R^d$ can be decomposed into its maximal connected sets, termed the connected components of $M$. We denote the set of connected components by $\conn(M)$. 

\begin{theorem} \label{fixPoint_upperBoundPWL}
Let $f$ be a multi-layer neural network as in (\ref{networkFunLayered}) with $\act$ a PWL activation function, where $0<k:=\vert\ar(\act)\vert-1$ and $ d_\nu:=\min\{\rank(\bW_{j}):j=1,\ldots,\nu\}$. Then
\begin{eqnarray*}
    \vert \afix(f)\vert\leq \vert \conn(\fix(f))\vert\leq \vert \ar(f)\vert \leq \prod_{\nu=1}^L \numb_{d_\nu}(n_\nu,k).
\end{eqnarray*} 
\end{theorem}
\begin{proof}
The proof is given in \cref{sec:Appendix_HyperProof}.
\end{proof}

Based on a construction of saw-tooth like function via a multi-layer ReLU network, cf. \cite{telgarsky2016benefits}, it can be shown that $\vert \afix(f)\vert$ can grow exponentially in the number layers:

\begin{theorem}\label{th:example_exponential}
There exists a ReLU network function\\ $f:\R\rightarrow\R$ such that $\vert \afix(f)\vert=\Omega(2^L)$, where $L$ is the number of layers.
\end{theorem}
\begin{proof}
    The constructions of such a network can be found in \cref{sec:Appendix_HyperProof}.
\end{proof}

For a PWL function in one variable, one observes that two distinct attractive fixed points cannot reside within a single linear region or in two adjacent linear regions. This observation readily extends to the multivariate case, showing that in \cref{fixPoint_upperBoundPWL}, strict inequality holds between the left-hand side and right-hand side.
\begin{proposition}\label{prop_noTwoAdjAFix}
Let $f$ be a PWL network function as in (\ref{networkFunLayered}) and let $\by_1,\by_2\in\afix(f)$, $\by_1\neq\by_2$ and both being contained in some linear region of $f$. Then $\conv(\by_1,\by_2)$ intersects at least three linear regions of $f$.
\end{proposition}
The proof is given in \cref{sec:Appendix_HyperProof}.

\section{Regions with Stable Fixed Points}\label{sec:nonlinearFP}

To show the utility of \cref{th:upperLowerPMHyper} for the case of a one-hidden-layer networks, we first consider $f(\bx)=\bW\Act(\bW^T\bx)$ as in (\ref{networkFun1}). Let $\act$ be a differentiable activation function. The Jacobian matrix of $f$ at a point $\bx \in \mathbb{R}^d$ is then given by $\Jac_f(\bx) = \bW \bD_\act(\bW^T\bx) \bW^T$, where $\bD_\act(\bW^T\bx)$ is a diagonal matrix with entries $\act^\prime((\bW^T\bx)^{(j)})$, $j=1,\ldots,n$, on its diagonal. It is evident that the Jacobian is symmetric and thus
\begin{equation}\label{jacobian_01}
\begin{split}
  \Vert\Jac_f(\bx)\Vert &= \max_{\|\mathbf{r}\|=1} \left|\mathbf{r}^T  \Jac_f(\bx) \mathbf{r}\right| \\
   &= \max_{\|\mathbf{r}\|=1} \left|\sum_{j=1}^n \act^\prime\left(\bom_j^T\bx\right) (\bom_j^T\mathbf{r})^2\right|,
\end{split}
\end{equation}
where $\bom_j$ are the columns of $\bW$.

Now, if $\bx^* \in \afix(f)$, then $\Vert J_f(\bx)\Vert<1$. It follows that the individual terms in the lower line of \cref{jacobian_01}, being non-negative, cannot be too large. Indeed, if for some $j \in \{1, \ldots, n\}$, we take $\br=\bom_j/\Vert\bom_j\Vert$ and $\act^\prime(\bom_j^T\bx)\Vert\bom_j\Vert^2 \geq 1$, then $\bx$ cannot be a stable fixed point. As for instance, let $\act$ be the tanh activation. Then $\vert\act^\prime(x) \vert \leq 1-(e^2-1)^2(e^2+1)^{-2}=:c\approx 0.42$ if, and only if, $\vert x\vert \geq 1$. Assuming $\|\bom_j\|^2 \geq c^{-1}$, we obtain the following necessary condition for $\bx$ to be a stable fixed point: For all $j \in \{1, \ldots, n\}$ either
\begin{equation} \label{hyper_sigmoid_1}
    \bom_j^T\bx \geq 1 \quad \text{or} \quad \bom_j^T\bx \leq -1.
\end{equation}

If we assume a PWL activation, such as $\act=$hard tanh, the regions defined by conditions as in (\ref{hyper_sigmoid_1}) can contain at most one attractive or stabel fixed point by \cref{fixedPointsPWL}. We recall that hard tanh is defined by $\act(x)=-1$ for $x<-1$, $\act(x)=x$ for $x\in[-1,1]$, and $\act(x)=1$ for $x>1$.

\begin{theorem}\label{th:exact_bound_attrfp}
Let $f:\R^d\rightarrow\R^d$ as in (\ref{networkFun00}) with $n\geq d$ and $\act=$hard tanh. Consider the following cases:
\begin{enumerate}
    \item\label{attrfp_cond1} Assume  $\bW=\bV^T$, and for the columns of $\bW$ denoted by $\bom_1,\ldots,\bom_n\in\R^d$, assume $\Vert \bom_j\Vert\geq 1.$
    
    \item\label{attrfp_cond2} Let the rows of $\bV$ be $\bnu_1,\ldots,\bnu_n\in\R^d$, and let $\bW=\bW_u \bsig \bW_v^T$ be the singular value decomposition, with singular values $\sigma_1\geq \ldots\geq\sigma_d$. Denote the columns of $\bW_v$ by $\bq_1,\ldots,\bq_d\in \R^n$. Assume that 
        \begin{equation}\label{cond:extact_fp_bound}
            \min_{j\in\{1,\ldots,n\}}\Vert \bnu_j\Vert\,\sum_{k=1}^d \sigma_k^2 \left(\bq_k^{(j)}\right)^2\geq 1.
        \end{equation}
\end{enumerate}
In both cases, (\ref{attrfp_cond1}) and (\ref{attrfp_cond2}), we have $\vert \afix(f)\vert \leq \numb_d(n,1)$.
\end{theorem}

\begin{proof}(\cref{th:exact_bound_attrfp})
Assume that (\ref{attrfp_cond2}) holds. We first also assume that $\bu=0$ in (\ref{networkFun00}). By \cref{lemm:lower_spJac}, the spectral norm of the Jacobian $\Jac_f(\bx)$ is lower bounded by 
\begin{equation}\label{rho_cond_th}
   \Vert\Jac_f(\bx)\Vert^2\geq \min_{j\in\{1,\ldots,n\}}\vert\act^\prime(\bnu_j^T\bx)\vert^2\,\Vert \bnu_j\Vert^2 \sum_{k=1}^d \sigma_k^2\left(\bq_k^{(j)}\right)^2.  
\end{equation}
Considering (\ref{cond:extact_fp_bound}), it thus follows that $\Vert\Jac_f(\bx)\Vert<1$ can only happen if $\act^\prime(\bnu_j^T\bx)=0$, which is gives the following necessary conditions for a stable fixed point of $f$:
\begin{equation}\label{cond_hardtanh}
    \bnu_j^T\bx\geq 1 \ \mathrm{or}\ \bnu_j^T\bx\leq -1, \ j\in\{1,\ldots,n\}.
\end{equation}

The latter holds only in the regions defined by $\poly^\pm(\cH)$, where $\cH$ is the arrangement that emerges from the hyperplanes defined by $\bnu_j^T\bx= 1$, $\bnu_j^T\bx= -1$, for $j \in \{1,\ldots,n\}$. 
The case that $\bu\neq 0$ holds in (\ref{networkFun00}) only amounts to a parallel shift of these hyperplanes.
By \cref{fixedPointsPWL}, $\vert C\cap \afix(f)\vert \leq 1$, and hence, the assertion follows from \cref{th:upperLowerPMHyper}.

If Condition (\ref{attrfp_cond1}) in \cref{th:exact_bound_attrfp} holds, the proof follows by similar arguments, wherein the identity in (\ref{jacobian_01}) is used in place of (\ref{rho_cond_th}).

\end{proof}


\section{Numerical experiments}\label{sec:numExp}

We conduct a brief analysis of the disparity between the upper bound presented in \cref{th:exact_bound_attrfp}, \cref{fixPoint_upperBoundPWL}, and an immediate upper bound derived from Zaslavsky's theorem, cf. (\ref{zaslavsky_upper}).

To this end, we compare upper bounds for $\vert\ar(f)\vert$, where $f(\bx) = \bW \Act\left(\bV\bx +\bu\right) +\bz$ is a one-hidden-layer network, as defined in (\ref{networkFun00}). Here, $\act$ is a PWL activation function with $\vert\ar(\act)\vert-1=k$.

This investigation amounts to comparing upper bounds for $\vert \poly(\cH)\vert$, where $\cH\in \cHP_d(n,k)$. According to \cref{fixPoint_upperBoundPWL}, $\numb_d(n,k)$ is an upper bound, while a direct application of Zaslavsky's theorem, cf. (\ref{zaslavsky_upper}), assuming the worst-case scenario where all hyperplanes are in general position, provides $\numb_d (n\cdot k)$ as an upper bound. Thus define:
\begin{equation}\label{def_gamma_ratio}
    \gamma(n,k,d):=\frac{\numb_d(n\cdot k)}{\numb_d(n, k)}.
\end{equation}
To compare the upper bounds in \cref{th:exact_bound_attrfp} and \cref{fixPoint_upperBoundPWL}, we assume that $f$ is a hard tanh network that fulfills the assumptions in \cref{th:exact_bound_attrfp}. We define:
\begin{equation}\label{def_eta_ratio}
    \eta(n,d):=\frac{\numb_d(n,1)}{\numb_d(n, 2)}.
\end{equation}

\begin{figure}[ht]
	\vskip 0.2in
	\begin{center}
            \centerline{\includegraphics[width=0.5\linewidth]{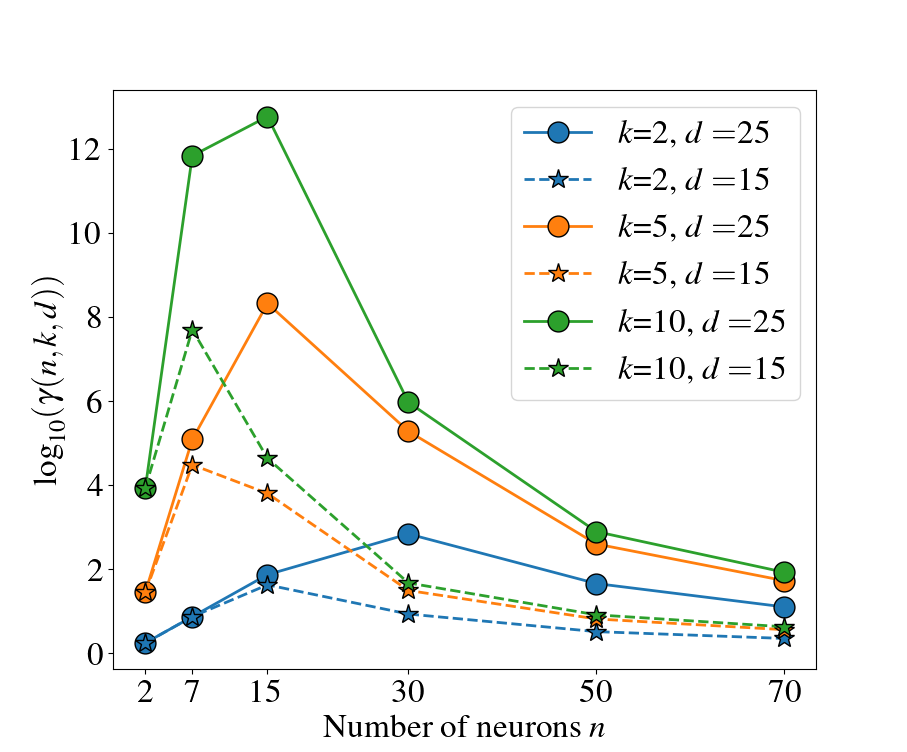}\hfill
           \includegraphics[width=0.5\linewidth]{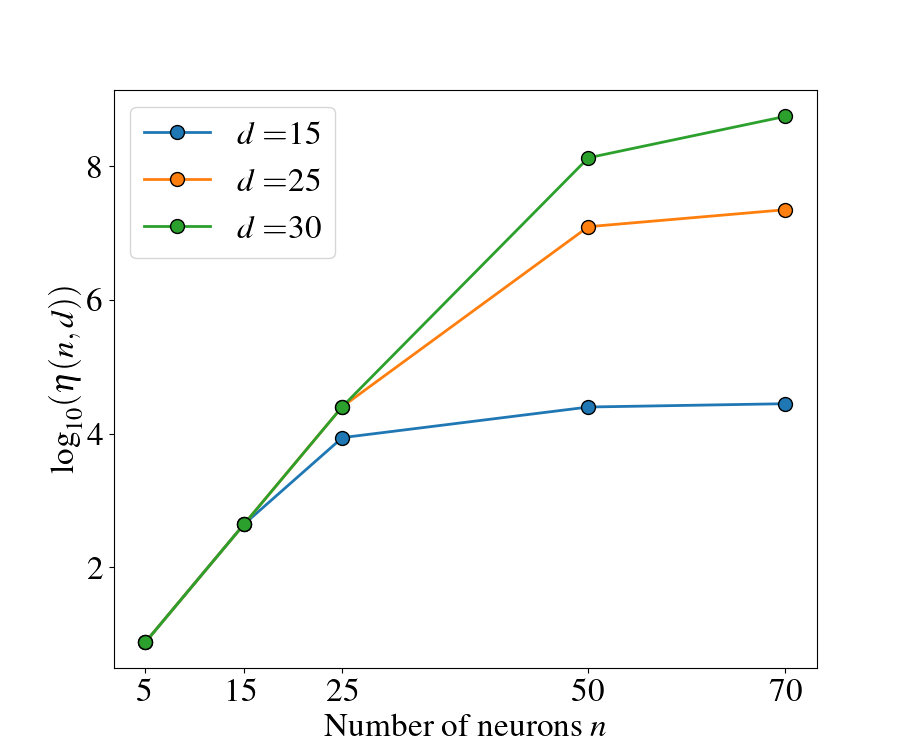}}
			
		\caption{\textbf{Left:} Plot of $\log_{10}(\gamma(n,k,d))$, as defined in (\ref{def_gamma_ratio}), with respect to the number of neurons $n$ for $\vert\ar(\act)\vert-1=k=2,5,10$ and $d=15,25$.
        \textbf{Right:} Plot of $\log_{10}(\eta(n,d))$, as defined in (\ref{def_eta_ratio}), with respect to the number of neurons $n$ and for $d=15,20,30$.} 
		\label{fig:upper_hyperplane_01}
	\end{center}
	\vskip -0.2in
\end{figure}

The evaluation of $\gamma(n,d,k)$ and $\eta(n,d)$ in \cref{fig:upper_hyperplane_01} reveals a significant improvement by several orders of magnitude.

The ratio of $\gamma(n,k)$ reaches its maximum in $[1,d]$, cf. \cref{fig:upper_hyperplane_01}. With the standard upper bound for (\ref{zaslavsky_upper}) for $n>d$, cf. Theorem 3.7 in \cite{anthony1999neural}, we have  $\gamma(n,k,d) \leq (enk/d)^d k^{-n}=:h(n)$ for $n\cdot k>d$ and $n\leq d$.
Taking derivatives reveals that $h$ reaches its maximum for $n=d/\ln(k)$. In our computational experiments, we observed that this serves as a good estimate for locating the maximum.

\section{Conclusion}\label{sec:concl}
 Upper bounds on the number of (stable) fixed points have been derived for networks with Piecewise Linear (PWL) activation functions, demonstrating improvement over bounds derived from existing results. Exploring whether such upper bounds can also be extended to the case of smooth activation functions remains a subject of future research.


\section{Auxiliary Results and Proofs}\label{sec:Appendix_HyperProof}

We slightly generalise the notation from \cref{def:paraHyperArr} for the subsequent proofs.
\begin{definition}\label{def:paraHyperArr_append}
    Let $H_{j,l}$ be hyperplanes in a $d-$dimensional Euclidean space, with $j\in[k_l]$ and $l=1,\ldots,n $, where $k_1,\ldots,k_n\in\N$ and where $[k]:=\{1,\ldots,k\}$ ($k\in\N$). For fixed $l$, all hyperplanes $H_{j,l}$ are \textbf{parallel but not equal}. We also assume that (\gputph) holds. The set of all arrangement as defined above are denoted by
    \begin{equation}
        \cHP_d((k_1,\ldots,k_n)).
    \end{equation}
    For the case that $k_1=\ldots k_n=:k$, we still briefly write $\cHP_d((k_1,\ldots,k_n))=\cHP_d(n,k)$ (as before).
\end{definition}

\begin{lemma} \label{lemma_parallelHs}
Let $\cH\in\cHP_d((k_1,\ldots,k_n))$ and $H_1, H_2$ be two parallel hyperplanes such that  $\{H_1, H_2\}\cup \cH\in\cHP_d((k_1,\ldots,k_n+2))$.
Then 
\begin{equation}\label{lemma_parallelHs_eq}
    \numb_d\left(\cH \cup \{H_1\}\right) =  \numb_d\left(\cH \cup \{H_2\} \right).
\end{equation}

\end{lemma}
Note that in \cref{lemma_parallelHs}, the condition $\{H_1, H_2\}\cup \cH\in\cHP_d((k_1,\ldots,k_n+2))$ entails that \gputph holds for $\{H_1, H_2\}\cup \cH$.
\begin{proof}(\cref{lemma_parallelHs})
     The assertion is obvious for $d=1$. To prove the result by induction over $d$, let us assume it holds true for $d-1\geq 0$.
     Let
     \begin{equation*}
    \begin{split}
    \cH^1&:= \{ H \cap H_1:H\in\cH,\, H\neq H_1,\, H\cap H_1\neq \emptyset \},\\
    \cH^2&:= \{ H \cap H_2:H\in\cH,\, H\neq H_2,\, H\cap H_2\neq \emptyset \}. 
    \end{split}
\end{equation*}
Then $\cH^1,\,\cH^2\in\cHP_{d-1}((k_1,\ldots,k_{n-1}))$ and by the induction hypothesis $\vert\poly(\cH^1)\vert=\vert\poly(\cH^2)\vert$. With Lemma 2.1 from \cite{rstandely2004}, 
\begin{equation}
   \numb_d\left(\cH \cup \{H_1\}\right)=\vert \poly(\cH)  \vert +\vert\poly(\cH^1)\vert= \vert \poly(\cH) \vert +\vert\poly(\cH^2)\vert =  \numb_d\left(\cH \cup \{H_2\}\right)
\end{equation}
and hence the identity in (\ref{lemma_parallelHs_eq}) is shown.
\end{proof}


\begin{lemma}\label{lemma_sameNumerRegionInHs}
Every arrangement $\cH\in\cHP_d((k_1,\ldots,k_n))$ has the same number of regions.   
\end{lemma}
\begin{proof}(\cref{lemma_sameNumerRegionInHs})
Every arrangement in
\begin{equation*}
    \cHP_d((\underbrace{1,\ldots,1}_{n\, \mathrm{times}}))
\end{equation*}
is in general position by definition. All of these arrangements thus have the same number of regions by Zaslavsky's theorem \cite{rstandely2004}, namely
\begin{equation}
  \numb_d((\underbrace{1,\ldots,1}_{n\, \mathrm{times}}))=\sum_{j=0}^d\binom{n}{j}.
\end{equation}
Thus, if take an arbitrary arrangement from this set and add a parallel hyperplane, say parallel to the one that corresponds to the first index, we obtain an arrangement 
\begin{equation*}
    \cH\in \cHP_d((2,\underbrace{1,\ldots,1}_{n-1\, \mathrm{times}}))
\end{equation*} 
and by Lemma 2.1 from \cite{rstandely2004}
\begin{equation*}
    |\poly(\cH)| = \numb_d((\underbrace{1,\ldots,1}_{n\, \mathrm{times}}))+\numb_{d-1}((\underbrace{1,\ldots,1}_{n-1\, \mathrm{times}})).
\end{equation*}
But this is independent of the concrete hyperplanes in the arrangements, so that iteratively, every arrangement $\cHP_d((k_1, \ldots, k_n))$ generates the same number of regions.
\end{proof}

\begin{lemma}\label{lemm:parallelHyper_recusion}
Let $n,k,d\in\N$, $d>1$, then
\begin{equation*}
    \numb_d(n,k)=\numb_d(n-1,k)+k\,\numb_{d-1}(n-1,k).
\end{equation*}
\end{lemma}

\begin{proof}
   By \cref{lemma_sameNumerRegionInHs}, every $\cH\in\cHP(n,k)$ has the same number of regions. Thus, consider some arbitrary $\cH$ in $\cHP(n,k)=\cHP_d((k_1,\ldots,k_n))$, where $k_j=k$ for $j=1,...,n$, cf. \cref{def:paraHyperArr_append}. Lemma 2.1 in \cite{rstandely2004} gives
   \begin{equation}\label{eq_parallelHyper_recusion}
     \numb_d(n,k)=\vert \poly(\cH)\vert=  \vert \poly(\cH\setminus\{H_{k,1}\})\vert+
     \vert\poly(\cH^{k,1})\vert
   \end{equation}
    where $\cH^{l,j}=\{H\cap H_{l,j}:H\in \cH\setminus\{H_{l,j}\}, H\cap H_{l,j}\neq \emptyset \}$, and where $H_{l,j}$ are the hyperplanes in $\cH$ with the indices as in \cref{def:paraHyperArr_append}. The $k$-fold iterative application of (\ref{eq_parallelHyper_recusion}) gives
    \begin{equation*}
     \numb_d(n,k)=\vert \poly(\cH)\vert=  \vert \poly(\cH\setminus\bigcup_{l=1}^k\{H_{l,1}\})\vert+
     \sum_{l=1}^k\vert\poly(\cH^{l,1})\vert=\numb_{d}(n-1,k)+k\, \numb_{d-1}(n-1,k).
   \end{equation*}
\end{proof}


\begin{lemma}\label{lemm:upperBoundHpyer_recusion}
     Let $\cH_{l,m}$ denote some arbitrary arrangement in $\cHP_m(l,2)$ as specified in \cref{def:hyperPlusMins}.
    For $d=1$, $\vert \poly^\pm(\cH_{n,1})\vert\leq n+1=:r(n,1)$, and for $n=1$, $\vert \poly^\pm(\cH_{1,d})\vert= 2=:r(1,d)$. Assume that for some $n,d>1$, we have $\vert \poly^\pm(\cH_{n-1,d})\vert\leq r(n-1,d) $ for all $\cH_{n-1,d}$, and $\vert \poly^\pm(\cH_{n-1,d-1})\vert\leq r(n-1,d-1) $ for all $\cH_{n-1,d-1}$. Then the following holds for all $\cH_{n,d}$:
    \begin{equation}\label{rec_upperBoundHpyer}
        \vert \poly^\pm(\cH_{n,d})\vert\leq r(n-1,d)+r(n-1,d-1)=:r(n,d).
    \end{equation}
    .
\end{lemma}

\begin{proof}
    As in the statement of \cref{lemm:upperBoundHpyer_recusion}, let $\cH_{l,m}\in\cHP_m(l,2)$ denote an arrangement as in \cref{def:hyperPlusMins}.
    
    The initial conditions for $d=1$, $\vert \poly^\pm(\cH_{n,1})\vert\leq n+1$, and for $n=1$, $\vert \poly^\pm(\cH_{1,d})\vert =2$ are obvious. 
    
    To verify (\ref{rec_upperBoundHpyer}), consider $\cH\in\cHP_d(n,2)$ and two parallel hyperplanes $H_1:=H_{1,1},\, H_2:=H_{2,1}$ (in the notation of \cref{def:hyperPlusMins}) in $\cH$ and set $\tilde{\cH}:=\cH\setminus\{H_1,H_2\} \in \cHP_d(n-1,2)$.
    We make the following observations.
    \begin{enumerate}
        \item \label{upper_rec_cond1} If $C\in\poly^\pm(\tilde{\cH})$ and $H_1$ intersects $C$, then $C$ is partitioned into two regions $C_1,C_2\in\poly(\tilde{\cH}\cup\{H_1\})$, exactly one of which belongs to $\poly^\pm(\tilde{\cH}\cup\{H_1\})$.
        \item \label{upper_rec_cond2} If $C\in\poly^\pm(\tilde{\cH}\cup\{H_1\})$ and $H_2$ intersects $C$, then $C$ is partitioned into two regions, $C_1,C_2 \in\poly(\cH)$, exactly one of which belongs to $\poly^\pm(\cH)$.
        \item \label{upper_rec_cond3} If $C\in \poly^\pm(\tilde{\cH})$, and both $H_1$ and $H_2$ do not intersect $C$, then $C\in\poly^\pm(\cH)$ if, and only if, $C$ is not located between $H_1,H_2$.    
        \item \label{upper_rec_cond4} If $C\notin \poly^\pm(\tilde{\cH})$, then, no matter the position of $H_1,H_2$, the regions that emerge after adding $H_1,H_2$, whether they intersect $C$ or not, do not belong to $\poly^\pm(\cH)$.  
    \end{enumerate}

    Now, let $H_1^+$ denote the half space (corresponding to $H_1$) defined by $\bv_1^T\bx >b_1$, and $H_2^-$ denote the half space (corresponding to $H_2$) defined by $\bv_1^T\bx <a_1$, c.f (\cref{def:hyperPlusMins}). We assume that $b_1$ is sufficiently small such that $H_1^+\cap C\neq \emptyset$ for all $C\in \poly(\tilde{\cH})$. Thus observation (\ref{upper_rec_cond1}) and (\ref{upper_rec_cond3}) above give
    \begin{equation}\label{upper_rec_eq1}
        \vert\poly^\pm(\tilde{\cH}\cup\{H_1\})\vert = \vert\poly^\pm(\tilde{\cH}) \vert \leq r(n-1,d).
    \end{equation}
    Note that observation (\ref{upper_rec_cond3}) applies, since $a_1<b_1$ and $b_1$ is chosen so small that $H_1^+\cap C\neq \emptyset$ for all $C\in \poly(\tilde{\cH})$, which implies that no $C\in \poly(\tilde{\cH})$ is located between $H_1,H_2$ when $H_2$ is added. Also, note that such a position of $H_1$ (with sufficiently small $b_1$) achieves the maximum possible number of regions in $\poly^\pm(\tilde{\cH}\cup\{H_1\})$ when $H_1$ is added to given $\tilde{\cH}$.

    Now, if $H_2$ is added to $\tilde{\cH}\cup\{H_1\}$, then according to observation (\ref{upper_rec_cond2}), $\vert\poly^\pm(\cH)\vert$ is maximized if $H_2$ intersects as many regions of $\poly^\pm(\tilde{\cH}\cup\{H_1\})$ as possible. This is equivalent to maximizing $\vert\poly^\pm(\cH^1)\vert$, where $\cH^1:=\{H\cap H_2:H\in\cH\setminus\{H_1,H_2\}\}$, and this is upper bounded by $r(n-1,d-1)$ by assumption. Adding the latter with (\ref{upper_rec_eq1}) yields (\ref{rec_upperBoundHpyer}). 
\end{proof}


\begin{lemma}\label{lemm:unfold_rec}
    For $n,d,k\in \mathbb{N}$, let $r(1,d):=k+1$, $r(n,1):=k\,n+1$, and for $n>1, d>1$
    \begin{equation}\label{recursion_def}
        r(n,d):=k\, r(n-1,d-1)+r(n-1,d).
    \end{equation}
    Then $r(n,d)$ unfolds to 
    \begin{equation}\label{recursion_unfold}
        r(n,d)=(k+1)\sum_{j=0}^{d-1}\binom{n-1}{j} k^j+k^{d}\sum_{j=d}^{n-1}\binom{j-1}{d-1}.
    \end{equation}
    For $k=1$, we have 
    \begin{equation}\label{recursion_unfold2}
        \sum_{j=0}^d\binom{n}{j}=2\sum_{j=0}^{d-1}\binom{n-1}{j} +\sum_{j=d}^{n-1}\binom{j-1}{d-1}.
    \end{equation}
\end{lemma}

Let us recall Pascal's rule for the following proof: $ \binom{n}{l} = \binom{n-1}{l-1} + \binom{n-1}{l}$

\begin{proof}(\cref{lemm:unfold_rec})
    It is immediately seen that (\ref{recursion_unfold}) fulfills the  initial conditions. For $n,d>1$, we obtain
    \begin{align*}
    r(n,d)&=k\, r(n-1,d-1)+r(n-1,d)\\
        &=k\,(k+1)\sum_{j=0}^{d-2}\binom{n-2}{j}k^j  +k\, k^{d-1}\sum_{j=d-1}^{n-2}\binom{j-1}{d-2} \\
        &+ (k+1)\sum_{j=0}^{d-1}\binom{n-2}{j} k^j      +k^d\sum_{j=d}^{n-2}\binom{j-1}{d-1}\\
        &=(k+1)\left(\sum_{j=1}^{d-1}\left(\binom{n-2}{j-1} +\binom{n-2}{j} \right)k^j+ \binom{n-2}{0}k^0\right)
        \\ &+k^d\left(\sum_{j=d-1}^{n-2}\binom{j-1}{d-2}+ \sum_{j=d}^{n-2}\binom{j-1}{d-1}\right)\\
        &=(k+1)\sum_{j=0}^{d-1}\binom{n-1}{j}k^j
        +k^d\left(\sum_{j=d-1}^{n-2}\binom{j-1}{d-2}+ \sum_{j=d}^{n-2}\binom{j-1}{d-1} \right) \\
        &=(k+1)\sum_{j=0}^{d-1}\binom{n-1}{j}k^j
        +k^d\left(\sum_{j=d}^{n-2}\binom{j}{d-1} +\binom{d-1}{d-1}\right)  \\
        &=(k+1)\sum_{j=0}^{d-1}\binom{n-1}{j}k^j
        +k^d\sum_{j=d}^{n-1}\binom{j-1}{d-1}.
    \end{align*}
    This shows that (\ref{recursion_def}) holds true for (\ref{recursion_unfold}).

    To verify (\ref{recursion_unfold2}) let $k=1$. Starting with the last line in the above equation, we have
    \begin{align*}
       2\sum_{j=0}^{d-1}\binom{n-1}{j}
        +\sum_{j=d}^{n-1}\binom{j-1}{d-1}
        &=\sum_{j=0}^{d-1}\binom{n}{j} +\binom{n-1}{d-1}+\sum_{j=d}^{n-1}\binom{j-1}{d-1}\\
        &=\sum_{j=0}^{d-1}\binom{n}{j} +\binom{n-1}{d-1}+\binom{n-1}{d}\\
        &=\sum_{j=0}^{d}\binom{n}{j},
    \end{align*}
    where Pascal's rule has been applied on consecutive summands in the left sum of the first line, and the identity $\sum_{l=0}^m\binom{n+l}{n}=\binom{n+m+1}{n+1}$ has been applied to obtain the second line.

\end{proof}

The terms in (\ref{recursion_unfold}) relate the recursion tree emerging from (\ref{recursion_def}) as follows:
\begin{equation*}
    (k+1)\sum_{j=0}^{d-1}\binom{n-1}{j} k^j
\end{equation*}
corresponds to all paths in that tree that end in a leaf with $n=1$, and
\begin{equation*}
    k^{d}\sum_{j=d}^{n-1}\binom{j-1}{d-1}
\end{equation*}
corresponds to all paths in that tree that end in a leaf with $d=1$ and $n>1$.

\begin{proof}(\cref{fixedPointsPWL})
Let's consider two points $\bx, \by \in C \cap \fix(f)$. Due to the convexity of $C$, $\conv(\bx, \by) \subset C$, implying that $f\vert_{\conv(\bx, \by)}$ is a linear (affine) function. Therefore, $h(t) = f(t\by + (1-t)\bx)$, $t \in [0,1]$, is also a linear (affine) function in the real variable $t$.
Since $h(0) = f(\bx) = \bx$ and $h(1) = f(\by) = \by$, we can conclude that $h$ is represented as $h(t) = \bx + t(\by - \bx)$. With this representation, we observe that $f(t\by + (1-t)\bx) = h(t) = \bx + t(\by - \bx)$ holds for all $t \in [0,1]$, demonstrating that $\conv(\bx, \by) \subset \fix(f)$.
\end{proof}


The next result, along with its accompanying proof, provides a more detailed insight into the role of hyperplane arrangements in PWL neural networks. In the following proof, we use a slight generalization in notation for linear regions: $\ar(f,\Omega):=\{C\cap M: C\in \ar(f)\}$, where $f$ represents a network function, and $M$ is a convex, linear manifold in the domain of $f$.

\begin{lemma} \label{lemma_intersectLinear}
Let $M\subset \R^n$ be a convex, linear manifold of dimension $d$, and $\act:\R\rightarrow\R$ be a PWL function with $1<k:=\vert\ar(\act)\vert-1$, and $\Act(\bx)=(\act(\bx^{(1)}),\ldots,\act(\bx^{(n)}))^T$, then 
\[
\vert\ar(\Act,M)\vert\leq \numb_d(n,k).
\]
\end{lemma}

\begin{proof}(Lemma \ref{lemma_intersectLinear})
We decompose $\Act(\bx)=(\act(\bx^{(1)}),\ldots,\act(\bx^{(n)}))^T$ as $\Act=h_n\circ h_{n-1}\circ \ldots\circ h_1$, where $h_j:\R^n \rightarrow \R^n$, $j=1,\ldots,n$, is defined by
\[\begin{pmatrix} \bx^{(1)}\\
\vdots\\
\bx^{(n)}
\end{pmatrix}
\mapsto
\begin{pmatrix}
\bx^{(1)}\\
\bx^{(2)}\\
\vdots\\
\bx^{(j-1)}\\
\act(\bx^{(j)})\\
\bx^{(j+1)}\\
\vdots\\
\bx^{(n)}
\end{pmatrix}.
\]
It is obvious that each $h_j$ is affine on $k$ subsets in $\R^n$, that are mutually separated by hyperplanes $\tilde{H}_{j,1},\ldots,\tilde{H}_{j,k}$, each of which is orthogonal to the $j-$th coordinate axis. Thus, $\Act$, as a function on $\R^n$, is linear (affine) on every region in 
\begin{equation}\label{upperBoundOneLayer}
    \poly\left(\left\{\tilde{H}_{jl}:j=1,\ldots,k,\  l=1,\ldots,n\right\}\right).    
\end{equation}
Now, for all $j=1,\ldots,k$ and all $l=1,\ldots,n$, we see that $H_{j,l}:=M\cap \tilde{H}_{j,l}$ is either a hyperplane in $M$, or is equal to $M$, or is empty. For the sake of an upper bound, we have to assume that $H_{j,l}$ for $j=1,\ldots,k$, $l=1,\ldots,n$ are hyperplanes in $M$. We can also assume that are \gputph holds, which is the generic case, and the corresponding number of regions upper bounds the other cases. Now, $\Act:M\rightarrow \R^n$ is linear (affine) on every region in 
\begin{equation}\label{upperBoundOneLayer1}
    \poly\left(\left\{H_{jl}:j=1,\ldots,k,\  l=1,\ldots,n\right\}\right). 
\end{equation}
The assertion follows, since the number of regions in (\ref{upperBoundOneLayer1}) is less or equal $\numb_d(n,k)$, where equality holds if the arrangement is \gputph, which is the generic case.
\end{proof}

\begin{proof}(\cref{fixPoint_upperBoundPWL})
Let us use $\A_j(\bx)=\bW_j\bx+\bu_j$, $j=1,\ldots,L$, to denote the linear (affine) mapping of layer $j$, cf. (\ref{networkFunLayered}). Recall that $d_\nu=\min\{\rank(\bW_j):j=1,\ldots,\nu\}$, $\nu=1,\ldots,L$.

Taking into account that $M_1=\A_1(\R^d)$ is a $d_1$-dimensional convex linear manifold (even an affine subspace) in $\R^{n_1}$, Lemma \ref{lemma_intersectLinear} yields
\begin{equation}\label{upBPWLeq1}
  \vert\ar(\Act\circ \A_1,\R^{d}) \vert\leq \vert\ar(\Act ,M_1) \vert \leq \numb_{d_1}(n_1,k).
\end{equation}
For some convex linear manifold $C\subset \ar(\Act ,M_1)\subset\R^{n_1}$ of dimension $d_1$ , the set $\A_2(C)$ is a $d_2$-dimensional convex linear manifold, say $M_2\subset \R^{n_2}$. Hence, Lemma \ref{lemma_intersectLinear} implies
\begin{equation}\label{upBPWLeq2}
    \vert\ar(\Act\circ \A_2,C) \vert\leq \vert\ar(\Act ,M_2) \vert \leq \numb_{d_2}(n_2,k).
\end{equation}
It is now observed that every $C\in \ar(\Act ,M_1)$ can at most be partitioned into $\numb_{d_2}(n_2,k)$ activation regions of $\Act\circ\A_2$. Hence, combining (\ref{upBPWLeq1}) and (\ref{upBPWLeq2}), we obtain
\begin{equation*}
    \vert\ar(\hi_2\circ \hi_1,\R^{d})\vert \leq  \numb_{d_1}(n_1,k)\, \numb_{d_2}(n_2,k).
\end{equation*}
Following these lines of argumentation to the last layer gives $\vert \ar(f)\vert \leq \prod_{\nu=1}^L \numb_{d_\nu}(n_\nu,k)$. Note that $\Act$ is also applied to the last layer, according to our convention in (\ref{networkFunLayered}). \cref{fixedPointsPWL} gives $\vert \conn(\fix(f))\vert\leq \vert \ar(f)\vert$, and $\vert \afix(f)\vert\leq \vert \conn(\fix(f))\vert$ follows by definition.
\end{proof}


\begin{proof}(\cref{prop_noTwoAdjAFix})
To establish the assertion by contradiction, let's assume $C_1, C_2 \in \ar(f)$ with $\by_1 \in C_1$, $\by_2 \in C_2$, and $\conv(\by_1,\by_2) \subset C_1 \cup C_2$. Since $f\vert_{\overline{C_1}}$ and $f\vert_{\overline{C_2}}$ are linear (affine), they have constant Jacobian matrices denoted by $W_{C_1}$ and $W_{C_2}$, respectively.
Since both, $\by_1$ and $\by_2$, are stable, we have $\Vert W_{C_1}\Vert <1$, $\Vert W_{C_2}\Vert <1$, according to our definition of the term in \cref{sec:prem}. 
(Recall that $\Vert \bA \Vert$ means the spectral norm of a matrix $\bA$.)

Next, $\conv(\by_1, \by_2) \subset C_1 \cup C_2$, implies that there exists $t^* \in [0,1]$ such that $\bx^* := t^*\by_1 + (1-t^*)\by_2 \in \overline{C_1} \cap \overline{C_2}$. Considering that $f(\by_1) = \by_1$ and $f(\by_2) = \by_2$, we obtain a contradiction:
\begin{eqnarray*}
    \Vert \by_1 - \by_2 \Vert &=& \Vert f(\by_1) - f(\by_2) \Vert \\
    &\leq& \Vert f(\by_1) - f(\bx^*) \Vert + \Vert f(\bx^*) - f(\by_2) \Vert \\
    &\leq& \Vert W_{C_1} \Vert \, \Vert \by_1 - \bx^* \Vert + \Vert W_{C_2} \Vert \, \Vert \bx^* - \by_2 \Vert \\
    &<& \Vert \by_1 - \bx^* \Vert + \Vert \bx^* - \by_2 \Vert = \Vert \by_1 - \by_2 \Vert.
\end{eqnarray*}
The last equality follows from the fact that $\by_1, \bx^*, \by_2$ are located on a straight line with $\bx^*$ between $\by_1$ and $\by_2$.
\end{proof}

\begin{lemma}\label{lemm:lower_spJac}
Let $f:\R^d\rightarrow\R^d$ as in (\ref{networkFun00}) with $\bu=0$, $n\geq d$ and $\bnu_1,\ldots,\bnu_n\in\R^d$ the rows of $\bV$. Let $\bW=\bW_u \bsig \bW_v^T$ be the singular value decomposition of $\bW$, with singular values $\sigma_1\geq \ldots\geq\sigma_d$ and $\bq_1,\ldots,\bq_d\in \R^n$ the columns of $\bW_v$. Then the spectral norm of the Jacobian $\Jac_f(\bx)$ is lower bounded as follows
\begin{equation*}
    \Vert\Jac_f(\bx)\Vert\geq \min_{j\in\{1,\ldots,n\}}\vert\act^\prime(\bnu_j^T\bx)\vert\,\Vert \bnu_j\Vert \sqrt{\sum_{k=1}^d \sigma_k^2 \left(\bq_k^{(j)}\right)^2}. 
\end{equation*}

\end{lemma}

\begin{proof}(\cref{lemm:lower_spJac})
    The Jacobian matrix of $f$ at $\bx\in\R^d$ is given by $\Jac_f(\bx) = \bW \bD_\act(\bV\bx)\bV$, where $\bD:=\bD_\act(\bV\bx)\in\R^{n\times n}$ is the diagonal matrix with entries $\act^\prime((\bV\bx)^{(j)})$ on its diagonal. Its spectral norm is thus given by
    \begin{equation}\label{jacobian_02}
        \begin{split}
         \Vert\Jac_f(\bx)\Vert^2 &= \max_{\|\mathbf{r}\|=1} \mathbf{r}^T (\Jac_f(\bx))^T \Jac_f(\bx) \mathbf{r}\\
           &= \max_{\|\mathbf{r}\|=1} \br^T\bV^T \bD\bW^T\bW \bD\bV\br   \\
           &= \max_{\|\mathbf{r}\|=1} \br^T\bV^T \bD\bW_v\bsig^2\bW_v^T \bD\bV\br   
        \end{split}
    \end{equation}
With $\br_j:=\bnu_j/\Vert \bnu_j\Vert$, and $\by_j:=\bD\bV\br_j$, we have 
\begin{equation}
      \Vert\Jac_f(\bx)\Vert^2\geq \sum_{k=1}^d \sigma_k^2 \vert \bq_k^T\by_j\vert^2
      \geq (\act^\prime(\bnu_j^T\bx))^2\Vert \bnu_j\Vert^2 \sum_{k=1}^d \sigma_k^2\left(\bq_k^{(j)}\right)^2.
\end{equation}

\end{proof}

Taking inspiration from \cite{telgarsky2016benefits}, we next construct at ReLU-network in a way that the number of stable fixed points increases exponentially in the number of layers.

\begin{proof}(\cref{th:example_exponential})
For some fixed $N\in\N$ let us define 
\begin{equation}\label{shiftedReluFun}
h_j(x)=\begin{cases}
\mathrm{ReLU}\left(x-j\frac{1}{2N}+\frac{1}{2N}\right) \ \ &\mathrm{if }\ j \ \mathrm{is \, odd} \\
\mathrm{ReLU}\left(x-(j-1)\frac{1}{2N}\right) \ \ &\mathrm{if }\ j \ \mathrm{is\, even},
\end{cases}
\end{equation}
for $j=1,\ldots,2N$.
And let $a_j$, $j=1,\ldots,2N$, be recursively defined by $a_1:=2N$ and
\begin{equation}\label{slopCoef1}
    a_j:=\begin{cases}
2N-\sum_{k=1}^{j-1}a_k \ \ &\mathrm{if }\ j>1 \ \mathrm{is\, odd} \\
-2N-\sum_{k=1}^{j-1}a_k \ \ &\mathrm{if }\ j \ \mathrm{is\, even}.
\end{cases}
\end{equation}
Then for 
\begin{equation}\label{saw_tooth_h}
    h=\sum_{j=1}^{2N}a_j h_j,
\end{equation}
one observes that, by the definition in (\ref{shiftedReluFun}), $h(x)=a_1h_1(x)$ for $x<\frac{1}{2N}$, $h(x)=a_1 h_1(x)+a_2 h_2(x)$ if $x<\frac{2}{2N}$, and in general $h(x)=a_1 h_1(x)+\ldots+a_j h_j(x)$ if $x<\frac{j}{2N}$. Further, the coefficients $a_j$ in (\ref{slopCoef1}) are arranged in a way that the slope of $h$ alternates between $\pm 2N$ and switches between the intervals $[0,1/2N]$,$[1/2N,2/2N]$,\ldots,$[(2N-1)/2N,1]$. Thus, the graph of $h$ describes a saw-tooth with 
\begin{equation}
    h(x)=0,\ \mathrm{for}\  x=\frac{j}{N},\, j=0,\ldots,N 
\end{equation}
and
\begin{equation}
 h(x)=1, \ \mathrm{for}\  x=\frac{j}{N}+\frac{1}{2N},\, j=0,\ldots,N-1. 
\end{equation}
Next, for some number of layers $L>1$, we define the $2N$ single neurons of the first layer by
\begin{equation*}
    \hi_1^{(k)}(x)=h_k(x), \ \mathrm{for }\  k=1,\ldots,2N.
\end{equation*}
For subsequent layers, let $\ba:=(a_1,\ldots,a_{2N})^T$ where the $a_1,\ldots a_{2N}$ are defined in (\ref{slopCoef1}), and let $\bz_j(x)=\hi_j\circ \ldots \circ \hi_1(x)$ denote the output at $j$-th layer. 
Then the $2N$ single neurons in such a layer are defined by
\begin{equation*}
    \hi_j^{(k)}(\bz_{j-1})=h_k(\ba^T \bz_{j-1} ), \ \mathrm{for }\ j=2,\ldots,L-1,\ k=1,\ldots,2N.
\end{equation*}
To follow the idea of the construction, let $\bA\in\R^{2N\times 2N}$ coincide row-wise with $\ba$. One verifies that $\bA\hi_1(x)$ coincides component wise with the saw-tooth function $h$ in (\ref{saw_tooth_h}). 
Then $h(x)$ passes $2M$ times through the interval $[0,1]$ as the argument $x$ runs once from left to right through $[0,1]$. Thus, $\bA\bz_2(x)$ is a saw-tooth function in each of its $2N$ components. The slope of these saw-tooth alternate between $\pm (2N)^2$ and switch between the intervals $[0,1/(2N)^2]$,\ldots,$[((2N)^2-1)/(2N)^2,1]$. In the similar way, it iteratively follows that $\bA\bz_j(x)$ is component wise a saw-tooth function taking values in $[0,1]$. The slope of these saw-tooth alternate between $\pm (2N)^j$ and that switch between the intervals $[0,1/(2N)^j]$,\ldots,$[((2N)^j-1)/(2N)^j,1]$. Thus the output $\bz_{L-1}$ has slope $\pm (2N)^{L-2}$, assuming that $L>1$. (Note that the slope would increase/decrease to $\pm (2N)^{L-1}$ for $\bA\, \bz_{L-1}$)

The last layer is arranged in a way that the steep slopes $(2N)^{L-2}$ produced by $\bz_{L-1}$ are reduced, which is needed to construct stable fixed points. To this end, we recursively define $c_j$, $j=1,\ldots,2N$, by $c_1:=(2 (2N)^{L-2})^{-1}$ and
\begin{equation}\label{slopCoef2}
    c_j:=\begin{cases}
\frac{1}{2(2N)^{L-2}}-\sum_{k=1}^{j-1}c_k \ \ &\mathrm{if }\ j>1 \ \mathrm{odd} \\
\frac{-1}{2(2N)^{L-2}}-\sum_{k=1}^{j-1}c_k \ \ &\mathrm{if }\ j \ \mathrm{even}.
\end{cases}
\end{equation}
With $\bc=(c_1,\ldots,c_N)^T$, we define $\bz_L(x):= \bc^T \bz_{L-1}(x)-1/2 \,(1/2N)^{L-1}$ and obtain a saw-tooth function that takes values in $[-1/2 \cdot 1/(2N)^{L-1},\, 1/2\cdot 1/(2N)^{L-1}]$, the slope of which alternates between $\pm 1/2$ switching between the intervals $[0,1/(2N)^{L-1}]$,\ldots,$[((2N)^{L-1} -1)/(2N)^{L-1},1]$. 
(Note that, in contrast to the multi-layer networks considered previously and introduced in (\ref{networkFunLayered}), the network constructed here applies only a linear activation to the last layer.)

Finally, to each layer we concatenate a residual connecting, each of which simply bypasses the input $x\in [0,1]$ and sums it with the saw-tooth function $\bz_L(x)$ described before. The whole network function thus writes as 
\begin{equation}
    f(x):=\bz_L(x)+x.
\end{equation}
This function has fixed points whenever $\bz_L(x)=0$, and in particular, has stable fixed points for all
\begin{equation*}
    x=\frac{2j-1}{(2N)^{L-1}}+\frac{1}{2(2N)^{L-1}},\ \ j=1,\ldots, \frac{(2N)^{L-1}}{2}.
\end{equation*}
That is, the points in $[0,1]$ where $\bz_L(x)=0$ and $\bz_L(x)$ has slope $-1/2$. 
\end{proof}

\bibliographystyle{plain} 
\bibliography{references}



\end{document}